\documentclass[runningheads]{llncs}
\usepackage{graphicx}
\usepackage{amsmath,amssymb} 
\usepackage{color}

\usepackage{epstopdf}
\usepackage{multirow}
\usepackage{authblk}
\usepackage[bookmarks=false]{hyperref}

\usepackage{longtable,multirow,hhline,tabularx,array,makecell,diagbox,colortbl,booktabs}

\usepackage{caption}
\usepackage{subfigure}

\begin{document}

\title{VIPL-HR: A Multi-modal Database for Pulse Estimation from Less-constrained Face Video} 
\titlerunning{VIPL-HR: A Less-constrained Video-based Pulse Estimation Database} 

\author{Xuesong Niu\inst{1,2} \and
Hu Han \inst{1,3,}\thanks{H. Han is the corresponding author.} \and
Shiguang Shan\inst{1,2,4} \and
Xilin Chen\inst{1,2}}
\authorrunning{X. Niu et al.} 

\institute{Key Laboratory of Intelligent Information Processing of Chinese Academy of Sciences (CAS), Institute of Computing Technology, CAS, Beijing 100190, China
\and University of Chinese Academy of Sciences, Beijing 100049, China \and
Peng Cheng Laboratory, Shenzhen, China \and
CAS Center for Excellence in Brain Science and Intelligence Technology, Shanghai, China \\
\email{xuesong.niu@vipl.ict.ac.cn; }
\email{\{hanhu, sgshan, xlchen\}@ict.ac.cn}}

%
%
%

\maketitle

\begin{abstract}
Heart rate (HR) is an important physiological signal that reflects the physical and emotional activities of humans. Traditional HR measurements are mainly based on contact monitors, which are inconvenient and may cause discomfort for the subjects. Recently, methods have been proposed for remote HR estimation from face videos. However, most of the existing methods focus on well-controlled scenarios, their generalization ability into less-constrained scenarios are not known. At the same time, lacking large-scale databases has limited the use of deep representation learning methods in remote HR estimation. In this paper, we introduce a large-scale multi-modal HR database (named as VIPL-HR), which contains 2,378 visible light videos (VIS) and 752 near-infrared (NIR) videos of 107 subjects. Our VIPL-HR database also contains various variations such as head movements, illumination variations, and acquisition device changes. We also learn a deep HR estimator (named as RhythmNet) with the proposed spatial-temporal representation, which achieves promising results on both the public-domain and our VIPL-HR HR estimation databases. We would like to put the VIPL-HR database into the public domain\footnote[1]{\url{http://vipl.ict.ac.cn/database.php}}.
\end{abstract}
\section{Introduction}

Heart rate (HR) is an important physiological signal that reflects the physical and emotional activities, and HR measurement can be useful for many applications, such as training aid, health monitoring, and nursing care. Traditional HR measurement usually relies on contact monitors, such as electrocardiograph (ECG) and contact photoplethysmography (cPPG), which are inconvenient for the users and limit the application scenarios. In recent years, a growing number of studies have been reported on remote HR estimation from face videos ~\cite{poh2010non,poh2011advancements,balakrishnan2013detecting,de2013robust,li2014remote,Tulyakov2016Self,wang2017algorithmic}.

\begin{figure}
\centering
\subfigure[]{
\includegraphics[width=0.45\linewidth, height=0.40\linewidth]{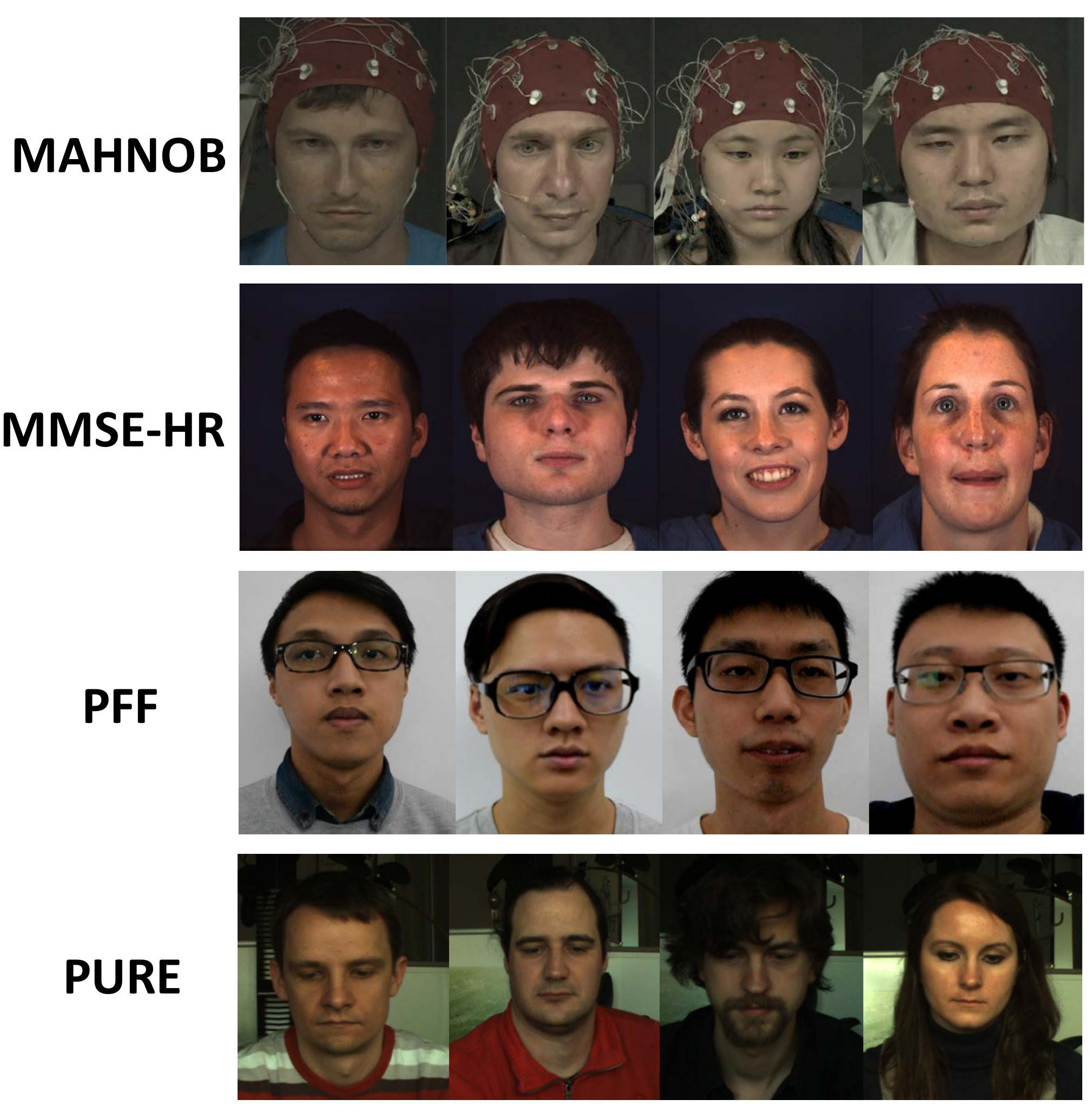}\vspace{2pt}
\label{fig:other_database}
}
\subfigure[]{
\includegraphics[width=0.45\linewidth, height=0.40\linewidth]{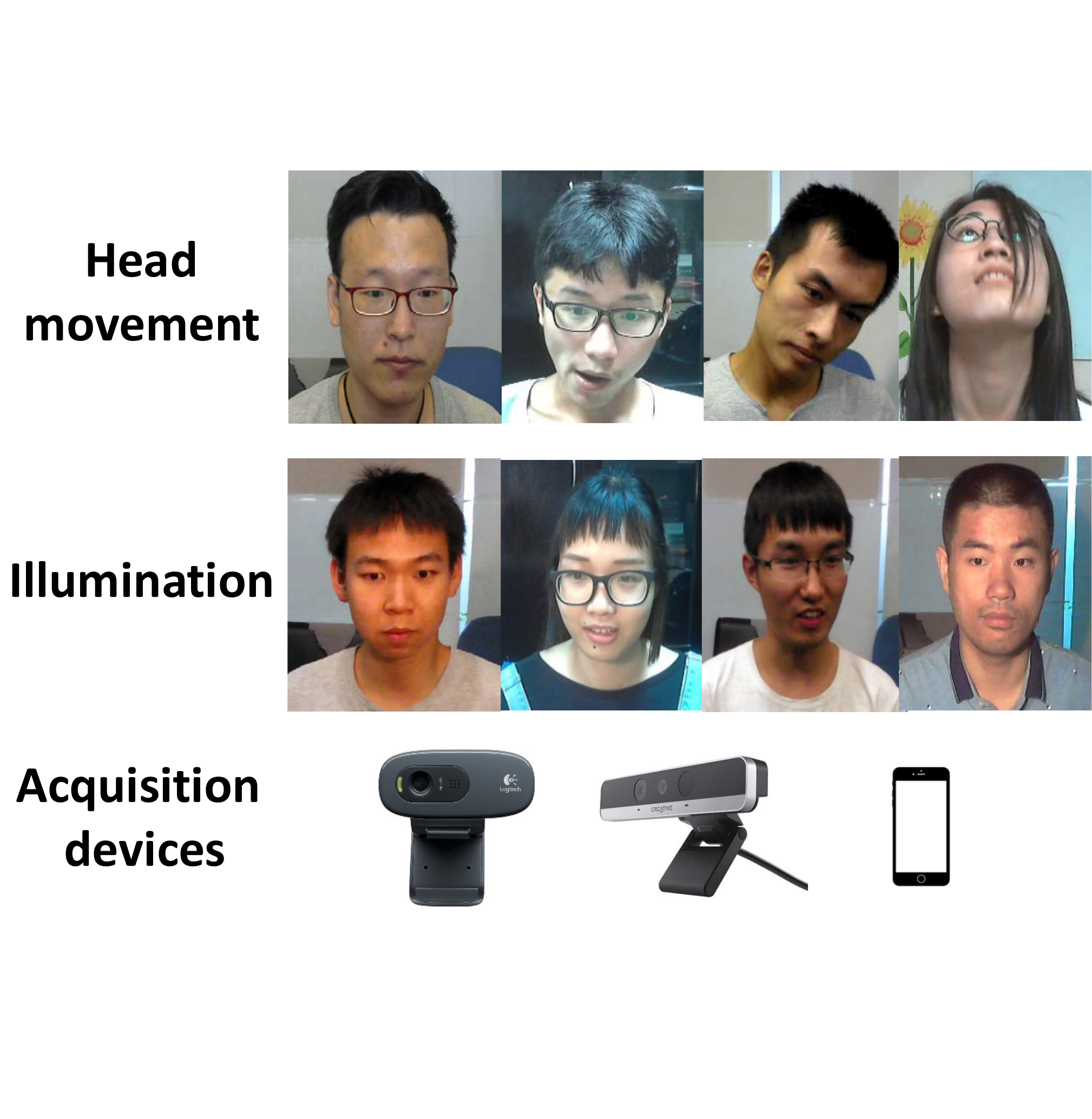}\vspace{2pt}
\label{fig:viplhr_database}
}
\caption{A comparison between (a) the public domain databases (MAHNOB, MMSE-HR, PFF, PURE) and (b) our VIPL-HR database in terms of the illumination condition, head movement, and acquisition device.}
\label{fig:other_vs_VIPLHR}
\end{figure}

The existing video-based HR estimation methods mainly depend on two kinds of signals: the remote photoplethysmography (rPPG) signals~\cite{poh2010non,poh2011advancements,de2013robust,li2014remote,Tulyakov2016Self,wang2017algorithmic} and the ballistocardiographic (BCG) signals~\cite{balakrishnan2013detecting}. There is no doubt that the idea of estimating HR from face videos could be a convenient physiological platform for clinical monitoring and health caring in the future. However, most of the existing approaches only provide evaluations on private databases, leading to difficulties in comparing different methods. Although a few public-domain HR databases are available~\cite{soleymani2012multimodal,Tulyakov2016Self,stricker2014non,hsu2017deep,li2018obf}, the sizes of these databases are very limited (usually smaller than 50 subjects). Moreover, these databases are usually captured in a well-controlled scenario, with minor illumination and motion variations (see Fig.~\ref{fig:other_database}). These limitations will lead to two issues for the remote HR estimation: i) it is hard to analyze the robustness of individual HR estimation algorithms against different variations and acquisition devices; ii) it is difficult to leverage the deep representation learning approaches in remote HR estimation, which are believed to have the ability to overcome the limitation of hand-crafted methods designed on specific assumptions~\cite{niu2018Synrhythm}.

To overcome these limitations, we introduce the VIPL-HR database for remote HR estimation, which is a large-scale multi-modal database recorded with various head movement, illumination variations, and acquisition device changes (see Fig.~\ref{fig:viplhr_database}). 

Given the large VIPL-HR database, we further propose a deep HR estimator, named as RhythmNet, for robust heart rate estimation from face. RhythmNet takes an informative spatial-temporal map as input and adopts an effective training strategy to learn the HR estimator. The results of within-database and cross-database experiments have shown the effectiveness of the proposed approach.

The rest of this paper is organized as follow: Section~\ref{related_work} discusses the related works of remote HR estimation and existing public-domain HR databases. Section~\ref{VIPL_HR_database} introduces the large-scale VIPL-HR database we have collected. Section~\ref{Proposed_Approach} provides the details of the proposed RhythmNet. After that, Section~\ref{experiments} evaluates the proposed method under both within-database and cross-database protocols. Finally, the conclusions and future work are summarized in Section~\ref{conculsion}.

\section{Related Work}
\label{related_work}

\subsection{Remote HR Estimation}

The possibility of using PPG signals captured by custom color cameras was firstly introduced by Verkruysse et.al~\cite{verkruysse2008remote}. Then many algorithms have been proposed, which can be generally divided into blind signal separation (BSS) methods, model-based methods and data-driven methods.

Poh et.al first applied the blind signal separation (BSS) to remote HR estimation~\cite{poh2010non,poh2011advancements}. They applied independent component analysis (ICA) to temporal filtered red, green, and blue (RGB) color channel signals to seek the heartbeat-related signal, which they assumed is one of the separated independent components. A patch-level HR signal calculation with ICA was performed later in~\cite{Lam2015Robust} and achieved the state-of-the-art on the public-available database MAHNOB-HCI~\cite{soleymani2012multimodal}.

Another kind of PPG-based HR estimation methods focus on leveraging the prior knowledge of the skin model to remote HR estimation. Haan and Jeanne firstly proposed a skin optical model of different color channels under the motion condition and computed a chrominance feature using the combination of RGB signals to reduce the motion noise~\cite{de2013robust}. In a later work of~\cite{wang2015exploiting}, pixel-wise chrominance features are computed and used for HR estimation. A detailed discussion of different skin optical model used for rPPG-based HR estimation is presented in~\cite{wang2017algorithmic}, and the authors further proposed a new projection method for the original RGB signals to extract pulse signals. In~\cite{niucontinuous}, Niu et al. further applied the chrominance feature~\cite{de2013robust} to continuous estimation situations.

Besides hand-crafted methods, there are also some data-driven methods designed for remote HR estimation. Tulyakov et.al~\cite{Tulyakov2016Self} divided the face into multiple regions of interest (ROI) to get a matrix temporal representation and used a matrix completion approach to purify rPPG signals. In~\cite{hsu2017deep}, Hsu et al. generated the time-frequency maps from different color signals and used them to learn an HR estimator. Although the existing data-driven approaches attempted to build learning based HR estimator, they failed to build an end-to-end estimator. Besides, the features they used remain hand-crafted, which may not be optimum for the HR estimation task. In~\cite{niu2018Synrhythm}, Niu et al. proposed a general-to-specific learning strategy to solve the problem of representing HR signals and lacking data. However, they didn't investigate the choice of color spaces and missing data situation.

Instead of the PPG-based HR measurement methods, the ballistocardiographic (BCG) signals, which is the subtle head motions caused by cardiovascular circulation, can also be used for remote HR estimation. Inspired by the Eulerian magnification method~\cite{wu2012eulerian}, Balakrishnan et al. tracked the key points on face and used PCA to get the pulse signal from the trajectories of feature points.~\cite{balakrishnan2013detecting}. Since these methods are based on subtle motion, the subjects' voluntary movements will introduce significant influence to the HR signals, leading to limited use in real-life applications.

Although the published methods have made a lot of progress in remote HR measurement, they still have limitations. First, the existing approaches are usually tested on well-controlled small-scale databases, which could not represent the real-life situations. Second, most of the existing approaches are designed in a step-by-step way using hand-crafted features, which are based on some specific assumptions and may fail in some complex conditions. Data-driven methods based on large-scale database are needed. However it is not as simple as visual attribute learning via deep neural network\cite{han2017heterogeneous}, in which the dominant information in the face image is related to the visual attribute learning task. The information in the face image related to the HR estimation task is quite weak.

\subsection{Public-domain Databases for Remote HR Estimation}

Many of the published methods reported their performance on private databases, leading to difficulties in performance comparison by other approaches. The first public domain database was introduced by Li et.al~\cite{li2014remote}. They evaluated their method on the MAHNOB-HCI database~\cite{soleymani2012multimodal}, which was designed for emotion detection and the subjects performed slight head movement and facial expressions. Later in 2016, Tulyakov et.al introduced a new database MMSE-HR~\cite{Tulyakov2016Self}, which was part of the MMSE database~\cite{zhang2016multimodal} and the subjects' facial expressions were more various. However, these two databases were originally designed for emotion analysis, and the subjects' motions were mainly limited to facial expression changes, which was far from enough for real-world remote HR estimation.

There are also a few public-available databases specially designed for the task of remote HR estimation. Stricker et al. firstly released the PURE database collected by the camera of a mobile sever robot~\cite{stricker2014non}. Hus et al. released the PFF database containing 10 subjects under 8 different situations~\cite{hsu2017deep}. These two databases are limited by the number of subjects and recording situations. In 2018, Xiaobai et al. proposed a database designed for HR and heart rate variability (HRV) measurement~\cite{li2018obf}. Since this database aims at HRV analysis and all the situations in this database are well-controlled, making it very easy for remote HR estimation.

The existing public-domain databases for remote HR estimation can be found in Table~\ref{table:database}. As we can see from Table~\ref{table:database}, all these databases are limited in either the number of subjects or the recording situations. A large-scale database recorded under real-life variations is required to push the studies on remote HR estimation.

{\setlength{\tabcolsep}{2mm}
\small
\begin{table}[t]
\begin{center}
\caption{A summary of the public-domain databases and VIPL-HR database.}
\label{table:database}
\begin{tabular}{cccccccc}
\hline\noalign{\smallskip}
\hline
  & \multirow{2}{*}{\textbf{Subject}} & \multirow{2}{*}{\textbf{Illumination}} & \textbf{Head} & \textbf{Recording} & \multirow{2}{*}{\textbf{Video}} \\
  & & & \textbf{movement} & \textbf{devices} & \\
\noalign{\smallskip}
\hline
\hline
\noalign{\smallskip}
MAHNOB~\cite{soleymani2012multimodal} & 27 &\emph{L}  & \emph{E} & \emph{C} & 527 \\
MMSE-HR~\cite{Tulyakov2016Self}	& 40 &\emph{L} 	& \emph{E} & \emph{C} & 102  \\
PURE~\cite{stricker2014non}  & 10 &\emph{L}	 & \emph{S}  & \emph{C} & 60  \\
PFF~\cite{hsu2017deep}	& 13 &\emph{L/D} & \emph{S/SM} & \emph{C} & 104  \\
OBF~\cite{li2018obf} & 106 &\emph{L}  &\emph{S} & \emph{C/N} & 2,120 \\
\noalign{\smallskip}
\hline
\hline
\noalign{\smallskip}
VIPL-HR & \textbf{107} & \textbf{L/D/B} & \textbf{S/LM/T} & \emph{\textbf{C/N/P}} & \textbf{3,130} \\
\hline
\end{tabular}
\smallskip
{\fontsize{8pt}\baselineskip\selectfont
\\ L = Lab Environment, D = Dark Environment, B = Bright Environment, E = Expression, \\ S = Stable, SM = Slight Motion, LM = Large Motion, T = Talking, \\ C = Color Camera, N = NIR Camera, P = Smart Phone Frontal Camera}
\end{center}
\end{table}
}

\section{The VIPL-HR Database}
\label{VIPL_HR_database}

In order to evaluate methods designed for real-world HR estimation, a database containing various face variations, such as head movements and illumination change, and acquisition diversity, is needed. To fill in this gap, we collected the VIPL-HR database, which contains more than one hundred subjects under various illumination conditions and different head movements. All the face videos are recorded by three different cameras and the relative physical measurements, such as HR, SpO2, and blood volume pulse (BVP) signal, are also simultaneously recorded. In this section, we introduce our VIPL-HR database from three aspects: i) setup and data collection, ii) video compression, and iii) database statistics.

\subsection{Setup and Data Collection}

We design our data collection procedure with two objectives in mind: i) videos should be recorded under natural conditions  (i.e., head movement and illumination change) instead of well-controlled situations; and ii) videos should be captured using various recording devices to replicate the common case in daily life, i.e., smartphones, RGB-D cameras, and web cameras. The recording setup is arranged based on these two targets, which includes a computer, an RGB web-camera, an RGB-D camera, a smartphone, a finger pulse oximeter, and a filament lamp. The details of the device specifications can be found in Table~\ref{table:devices}.

Videos recorded from different devices are the core component of VIPL-HR database. In order to test the influence of cameras with different recording quality, we choose the widely used web-camera Logitech C310 and the color camera of RealSense F200 to record the RGB videos. At the same time, while smartphones have become an indispensable part of our daily lives, remote HR estimation from the videos recorded by smart phone cameras has not been studied yet. Thus, we use a HUAWEI P9 smart phone (with its frontal camera) to record the RGB face videos for the potential applications of remote HR estimation on mobile devices. Besides recording the RGB color videos, we also record the NIR face videos using a RealSense F200 to investigate the possibility of remote HR estimation under dim lighting conditions. Related physiological signals, including HR, SpO2, and BVP signals, are synchronously recorded with a CONTEC CMS60C BVP sensor.
{
\setlength{\tabcolsep}{4pt}
\begin{table*}[t]
\begin{center}
\caption{Specifications and settings of individual recording devices used in our VIPL-HR database.}
\label{table:devices}
\begin{tabular}{cccc}
\hline\noalign{\smallskip}
\hline
\textbf{Device} & \textbf{Specification} & \textbf{Setting}  & \textbf{Output}\\
\noalign{\smallskip}
\hline
\hline
\noalign{\smallskip}
Computer & Lenovo ThinkCentre & Windows 10 OS & N/A\\
\hline
\noalign{\smallskip}
\multirow{2}{*}{Color camera} &  \multirow{2}{*}{Logitech C310} & 25fps & \multirow{2}{*}{Color videos}\\
&& 960$\times$720 color camera &\\
\hline
\noalign{\smallskip}
\multirow{2}{*}{RGB-D camera} & \multirow{2}{*}{RealSense F200} & 30fps, 640$\times$480 NIR camera & Color videos \\
 & & 1920$\times$1080 color camera, & NIR videos \\
\hline
\noalign{\smallskip}
\multirow{2}{*}{Smart phone}	& HUAWEI P9 & 30fps,  & \multirow{2}{*}{Color videos}\\
 &  frontal camera & 1920$\times$1080 color camera & \\
\hline
\noalign{\smallskip}
\multirow{2}{*}{BVP recoder} & \multirow{2}{*}{CONTEC CMS60C} & \multirow{2}{*}{N/A} & HR, SpO2, \\
& & & and BVP signals \\
\hline
\noalign{\smallskip}
Filament lamp & N/A & 50Hz & N/A \\
\hline
\end{tabular}
\end{center}
\end{table*}
}

\begin{figure}[b]
\centering
\includegraphics[width=0.6\linewidth]{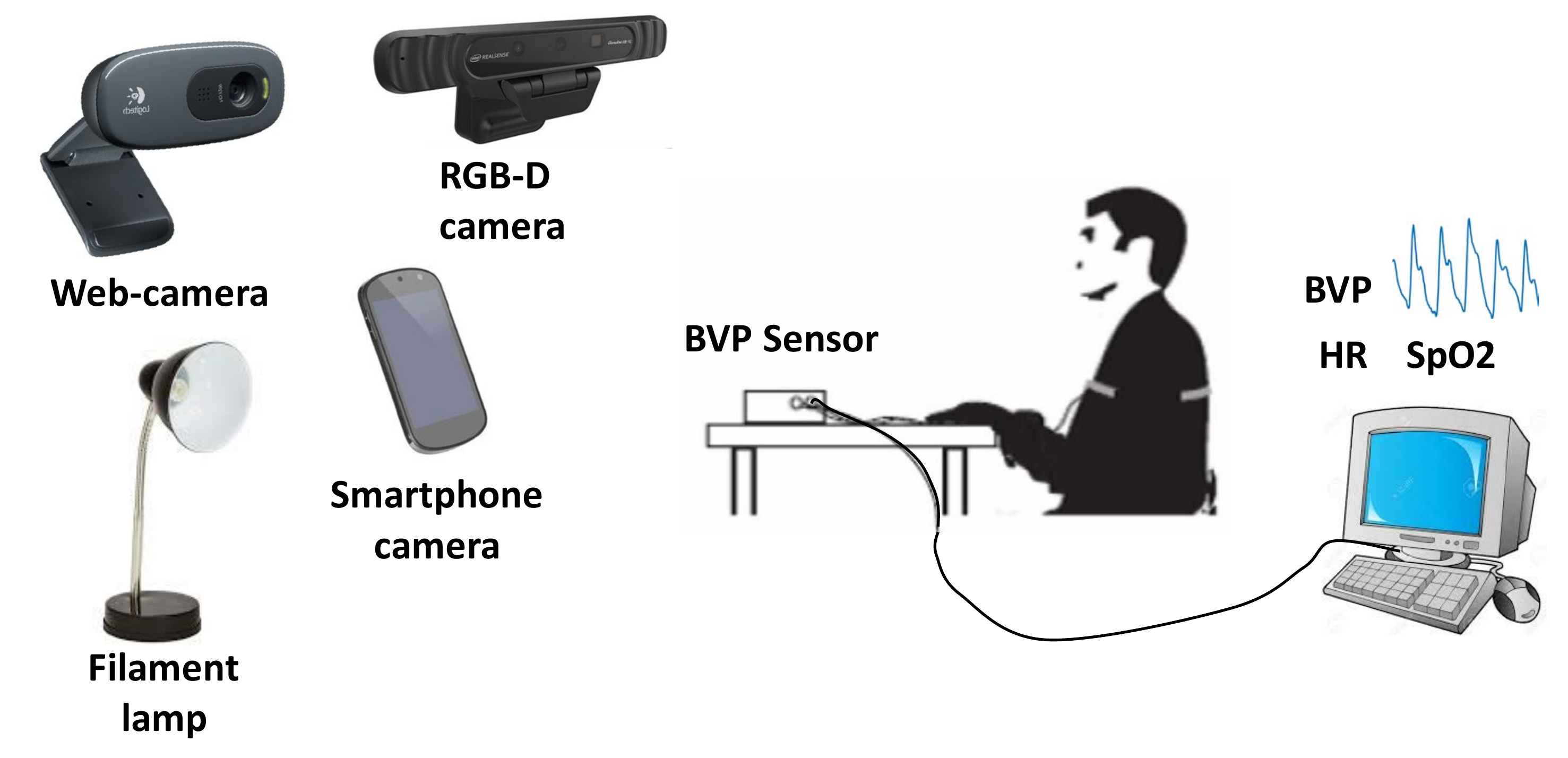}
\caption{An illumination of the device setup used in collecting the VIPL-HR database.}
\label{fig:setup}
\end{figure}

The recording environmental setup is illustrated in Figure~\ref{fig:setup}. The subjects are asked to sit in front of the cameras at two different distances: one meter and 1.5 meters. A filament lamp is placed aside the cameras to change the light conditions. Each subject is asked to sit naturally in front of the cameras, and daily activities such as talking and looking around are encouraged during the video recording. HR changes of the subject after exercises are also taken into consideration. The smartphone is first fixed in front of the subject for video recording, and then we asked the subject to hold the smartphone by themselves to record videos like a video chat scenario. Videos under nine different situations are recorded in total for each subject, and the details of these situations are listed in Table~\ref{table:record_situation}.

\setlength{\tabcolsep}{4pt}
\begin{table*}[t]
\begin{center}
\caption{Details of the nine recording situations in the VIPL-HR database.}
\label{table:record_situation}
\begin{tabular}{cccccc}
\hline\noalign{\smallskip}
\hline
\multirow{2}{*}{\textbf{Situation}} & \textbf{Head} & \multirow{2}{*}{\textbf{Illumination}} & \multirow{2}{*}{\textbf{Distance}} & \multirow{2}{*}{\textbf{Exercise}} & \textbf{Phone}\\
 & \textbf{movement} & & & & \textbf{position} \\
\noalign{\smallskip}
\hline
\hline
\noalign{\smallskip}
1 & S & L & 1m & No & Fixed \\
2 & LM & L & 1m & No & Fixed \\
3 & T & L & 1m & No & Fixed\\
4 & S & B & 1m & No & Fixed\\
5 & S & D& 1m & No & Fixed \\
6 & S & L & 1.5m & No & Fixed\\
7 & S & L & 1m & Yes & Fixed \\
8 & S & L & - & No & Hand-hold \\
9 & LM & L & - & No & Hand-hold \\
\noalign{\smallskip}
\hline
\end{tabular}
{
\\ S = Stable, LM = Large Motion, T = Talking, \\ L = Lab Environment, D = Dark Environment, B = Bright Environment}
\end{center}
\end{table*}

\subsection{Database Compression}
As stated in~\cite{mcduff2017impact}, video compression plays an important role in video-based heart rate estimation. The raw data of VIPL-HR we collected is nearly 1.05TB in total, making it very inconvenient for the public access. In order to balance the convenience of data sharing and completeness of HR signals, we investigate to make a compressed and resized version of our database, which can retain the completeness of the HR signals as much as possible. The compression methods we considered include video compression and frame resizing. The video compression codecs we take into consideration are `MJPG', `FMP4', `DIVX', `PIM1' and `X264', which are commonly-used video compression codecs. The resizing scales we consider are 1/2, 2/3, and 3/4 for each dimension of the original frame. We choose one of the widely used remote HR estimation method Haan2013~\cite{de2013robust} as a baseline HR estimation method to verify the HR estimation accuracy changes after individual comparison approaches.

\begin{figure}[t]
\centering
\subfigure[]{
\includegraphics[width=0.45\linewidth, height=0.22\linewidth]{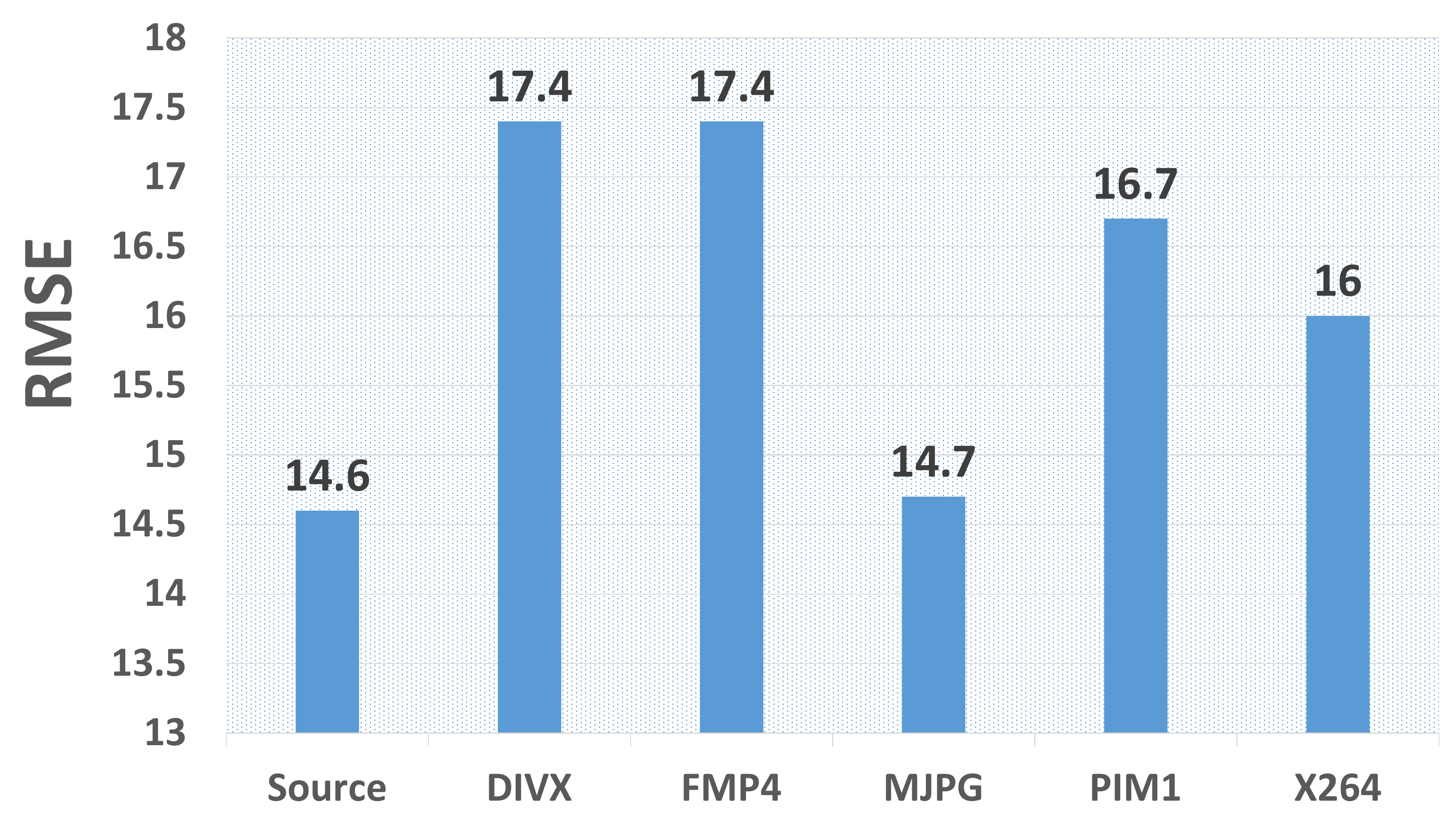}
}
\subfigure[]{
\includegraphics[width=0.45\linewidth, height=0.22\linewidth]{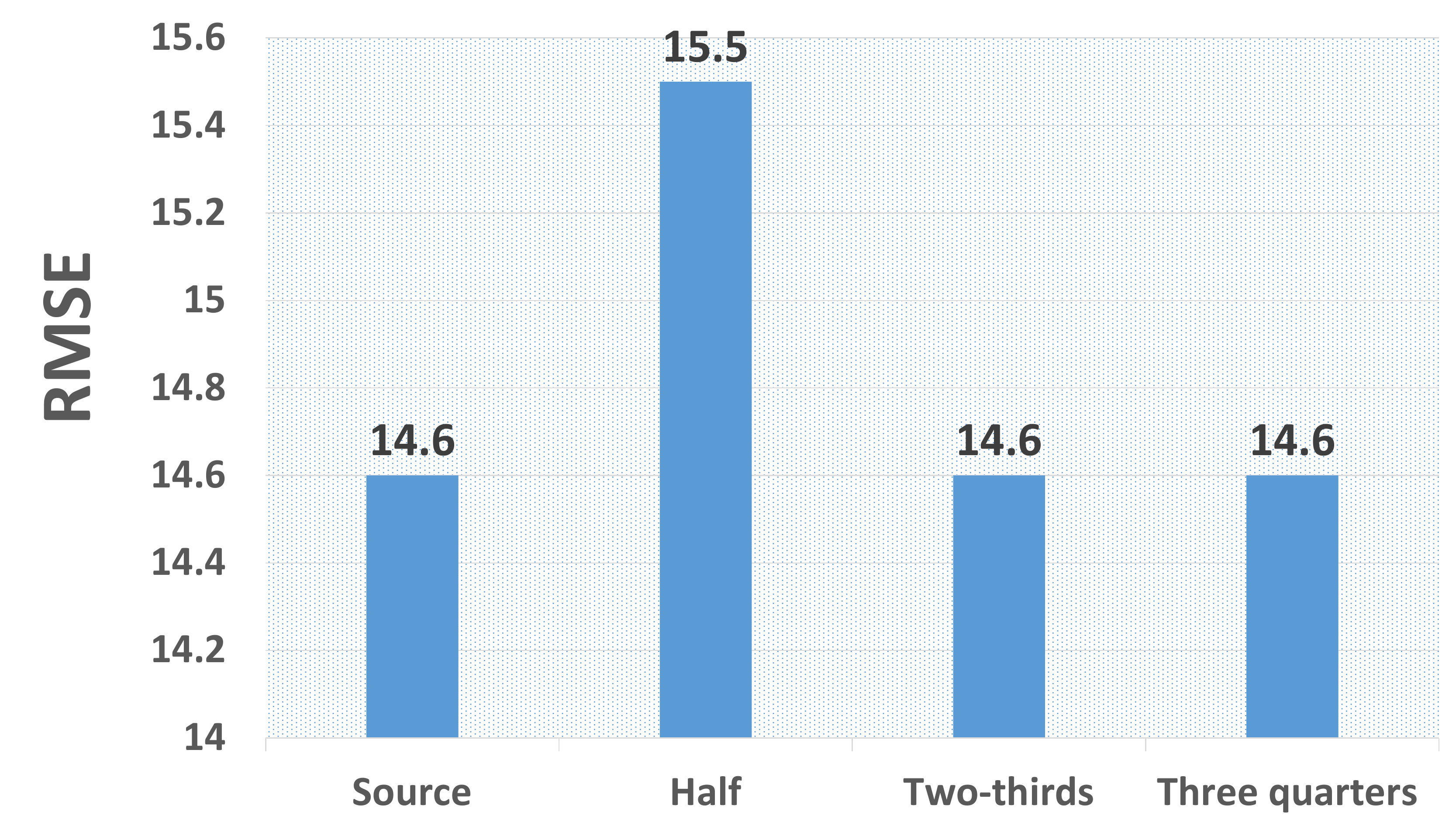}
}
\caption{Evaluations of using (a) different video compression codecs and (b) different image resolutions to compress the VIPL-HR database. All the compression methods are tested using Haan2013~\cite{de2013robust}. Source represents the test results on uncompressed data.}
\label{fig:video_compression}
\end{figure}

The HR estimation accuracies by the baseline HR estimator on various compressed videos are given in Fig.~\ref{fig:video_compression} in terms of root mean square error (RMSE). From the results, we can see that the `MJPG' video codec is better in maintaining the HR signal in the videos while it is able to reduce the size of the database significantly. Resizing the frames to two-thirds of the image resolution leads to little damage to the HR signal. Therefore, we choose the `MJPG' codec and two-thirds of the original resolution as our final data compression solution, and we obtained a compressed VIPL-HR dataset with about 48GB. However, we would like to share both the uncompressed and compressed databases to the research community based on the researchers' preference.

\subsection{Database Statistics}

\begin{figure}[t]
\centering
\subfigure[]{
\includegraphics[width=0.16\linewidth, height=0.16\linewidth]{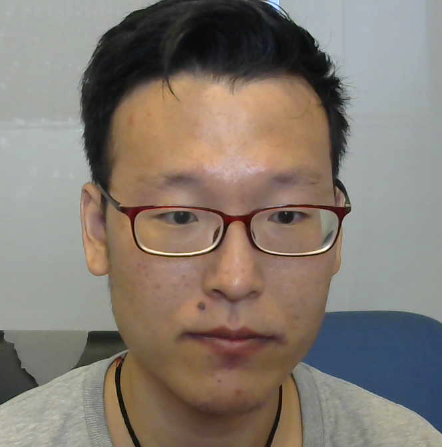}\vspace{2pt}
}
\subfigure[]{
\includegraphics[width=0.16\linewidth, height=0.16\linewidth]{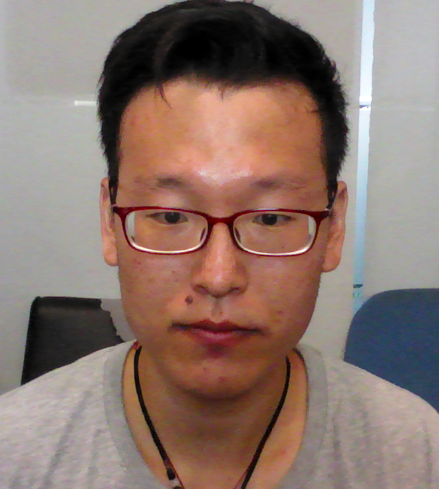}\vspace{2pt}
}
\subfigure[]{
\includegraphics[width=0.16\linewidth, height=0.16\linewidth]{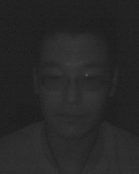}\vspace{2pt}
}
\subfigure[]{
\includegraphics[width=0.16\linewidth, height=0.16\linewidth]{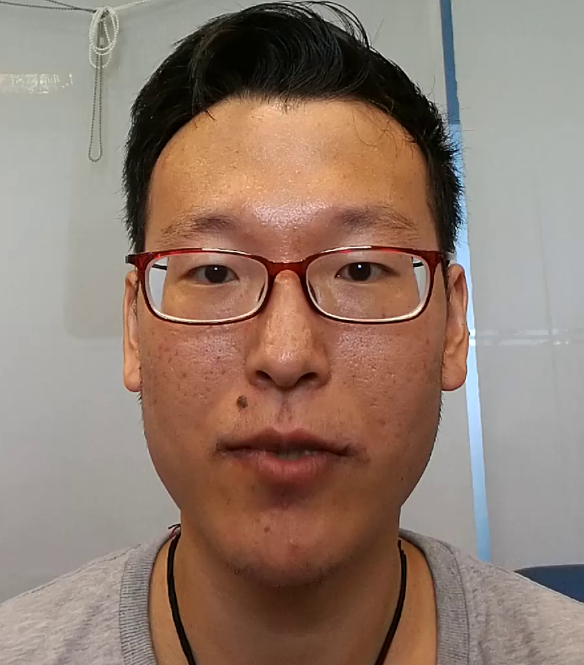}\vspace{2pt}
}
\caption{Example video frames captured for one subject by different devices: (a) Logitech C310, (b) RealSense F200 color camera, (c) RealSense F200 NIR camera, (4) HUAWEI P9 frontal camera.}
\label{fig:camera_image}
\end{figure}

The VIPL-HR dataset contains a total of 2,378 color videos and 752 NIR videos from 107 participants (79 males and 28 females) aged between 22 and 41. Each video is recorded with a length of about 30s, and the frame rate is about 30fps (see Table~\ref{table:database}). Some example video frames of one subject captured by different devices are shown in Fig.~\ref{fig:camera_image}.

\begin{figure}[t]
\centering
\subfigure{
\includegraphics[width=0.31\linewidth, height=0.20\linewidth]{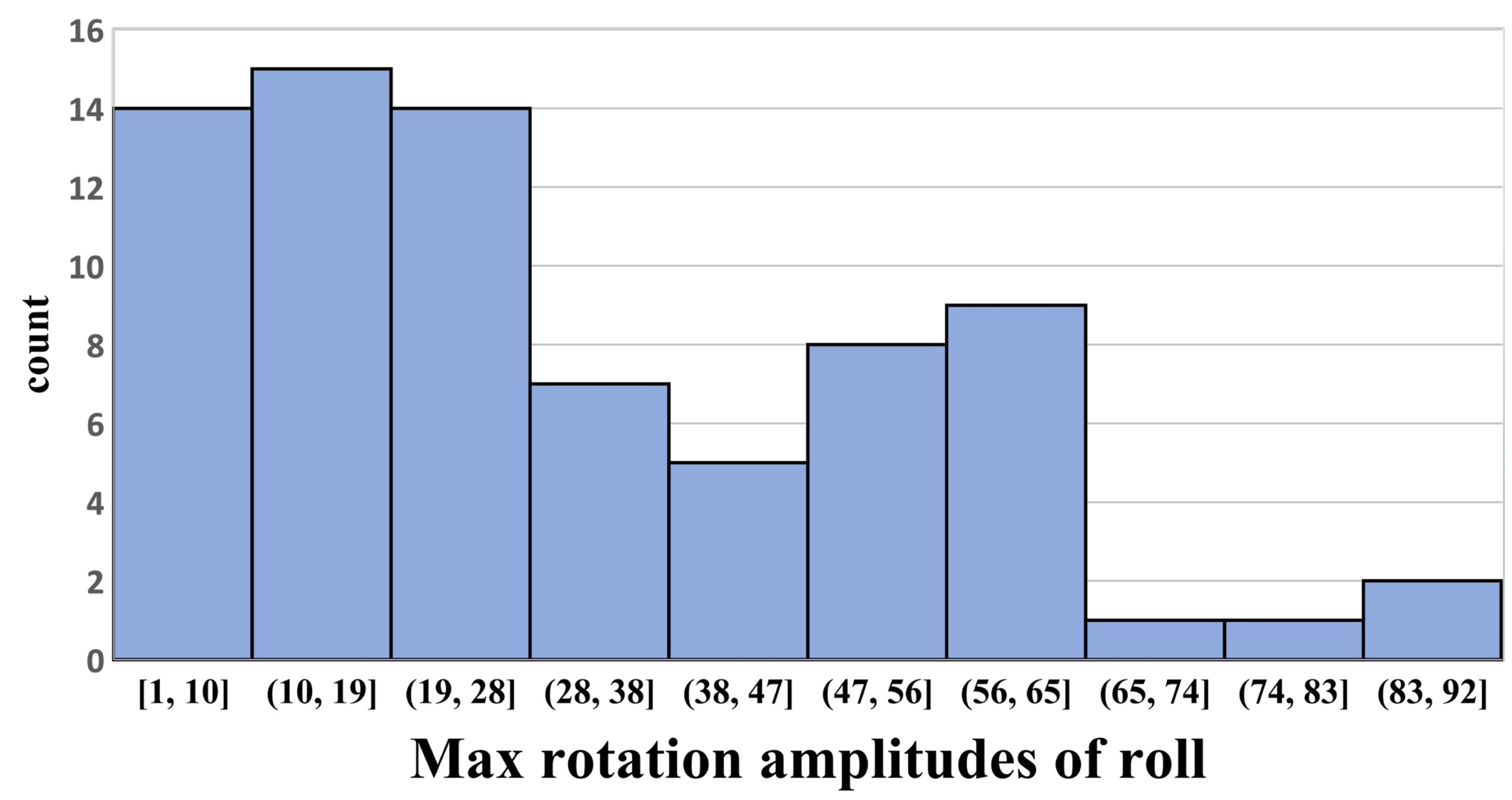}
}
\subfigure{
\includegraphics[width=0.31\linewidth, height=0.20\linewidth]{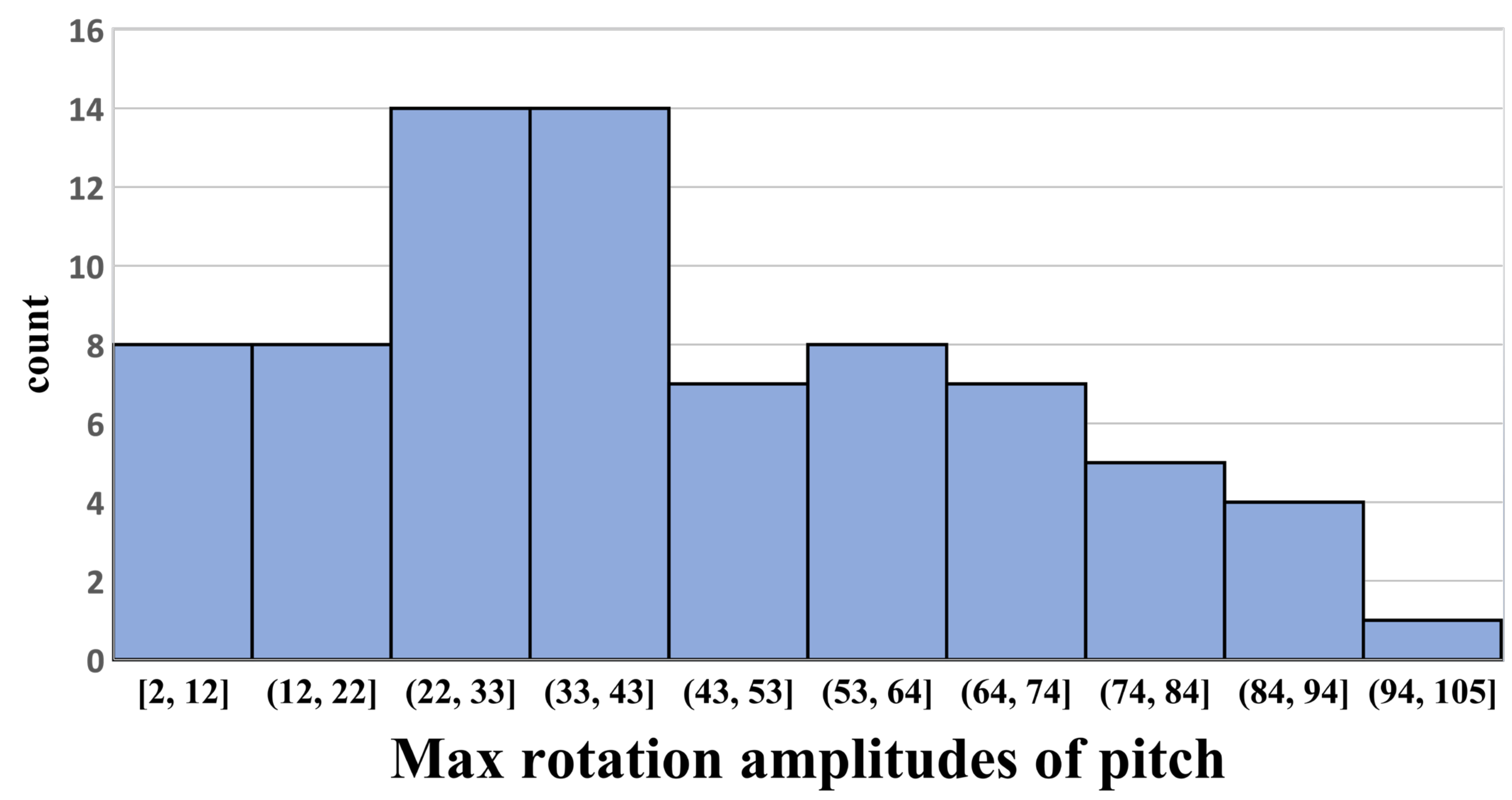}
}
\subfigure{
\includegraphics[width=0.31\linewidth, height=0.20\linewidth]{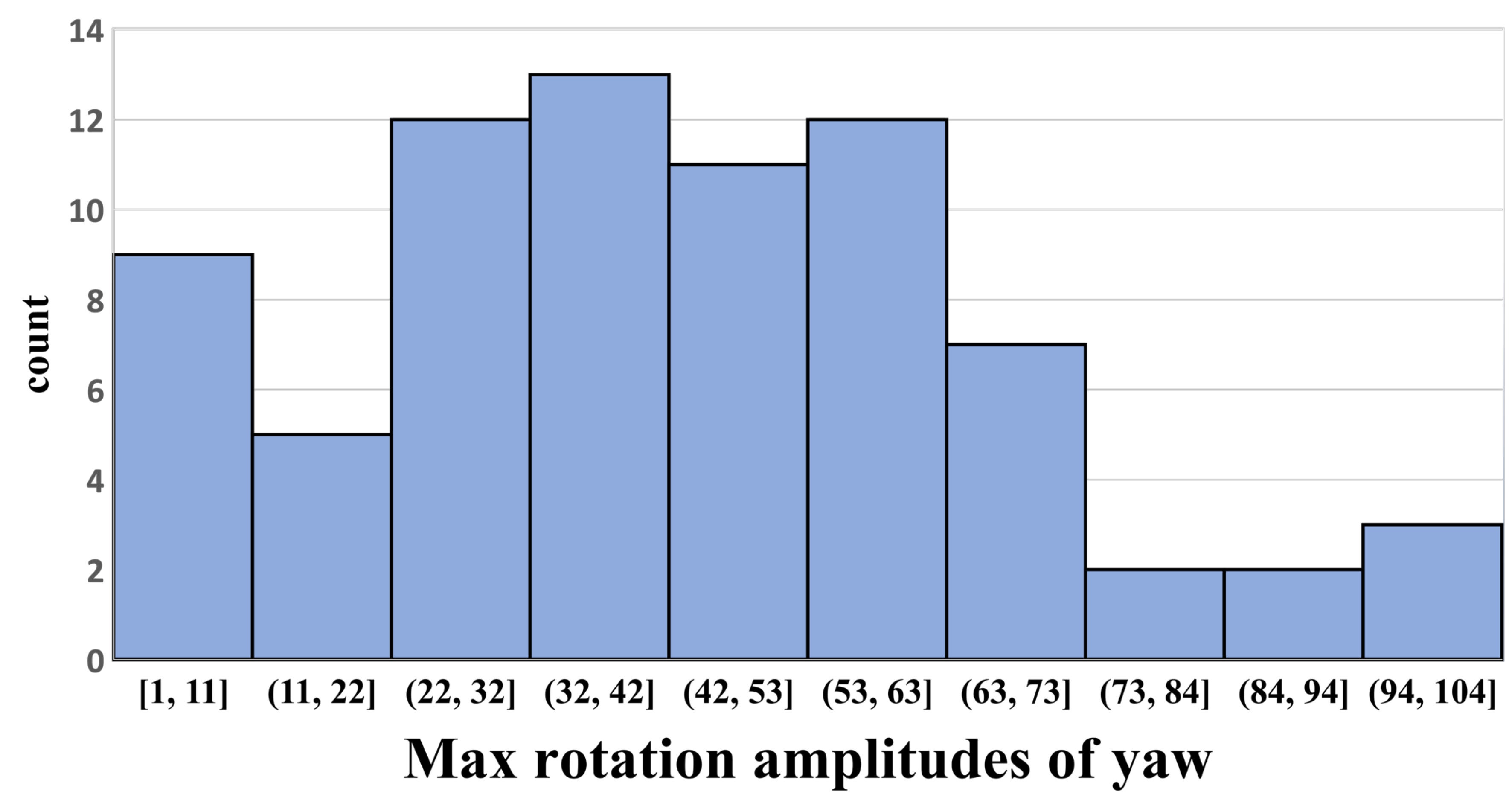}
}
\caption{ Histograms for maximum amplitudes of the roll, pitch and yaw rotation components for all the videos with head movement.}
\label{fig:head_motion}
\end{figure}

To further analyze the characteristics of our VIPL-HR database, we calculated the head pose variations using the OpenFace head pose estimator\footnote[2]{\url{https://github.com/TadasBaltrusaitis/OpenFace}} for the videos with head movement (see Situation 2 in Table~\ref{table:record_situation}). Histograms for maximum amplitudes of the three rotation components for all the videos can be found in Fig.~\ref{fig:head_motion}. From the histograms, we can see that the maximum rotation amplitudes of the subjects vary in a large range, i.e., the maximum rotation amplitudes are $92^\circ$ in roll, $105^\circ$ in pitch and $104^\circ$ in yaw. This is reasonable because every subject is allowed to look around during the video recording.

\begin{figure}
\centering
\includegraphics[width=0.6\linewidth, height = 0.23\linewidth]{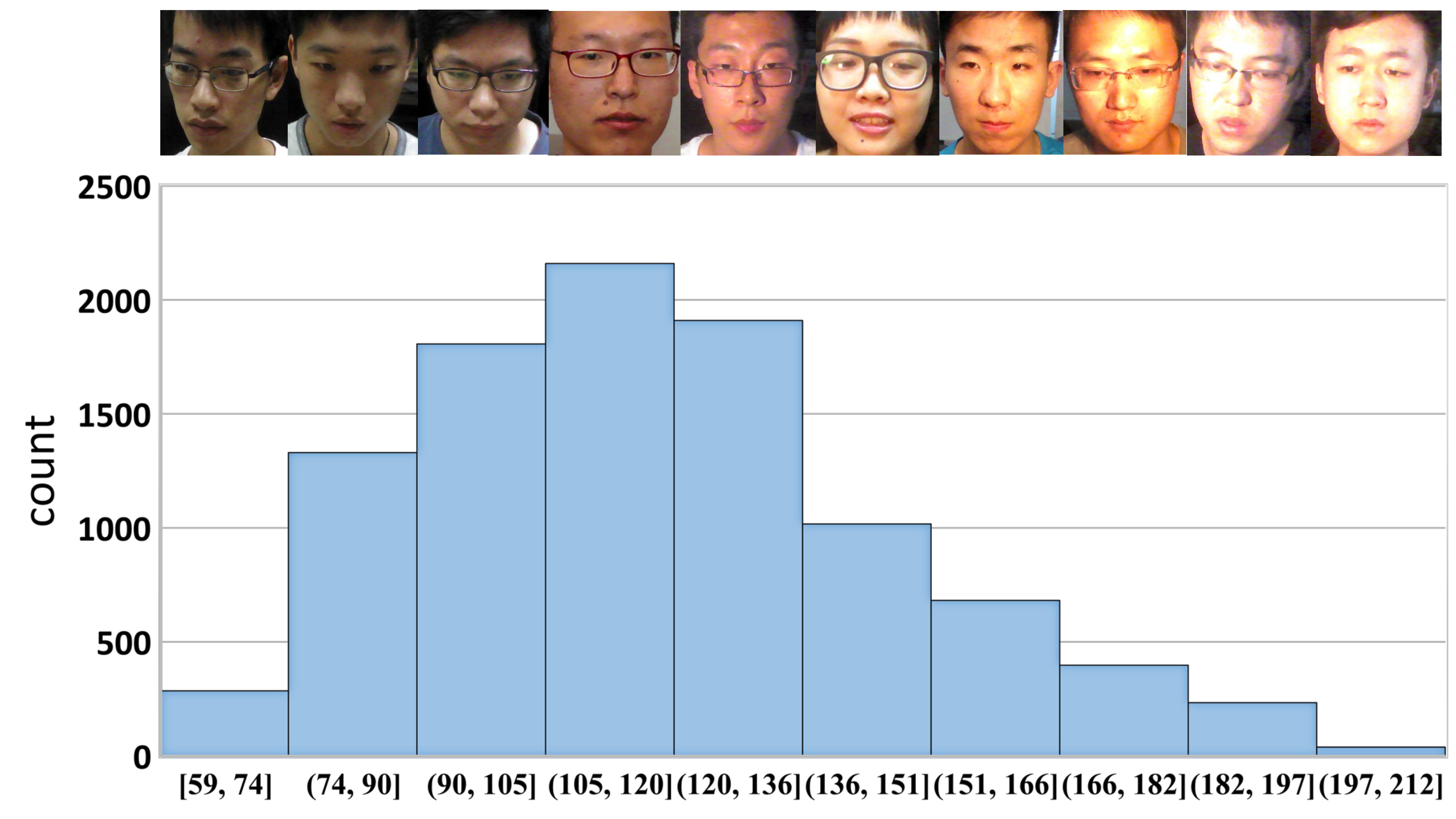}
\caption{ A histogram of the average image intensity (gray-scale) for the videos recorded under the illumination-specific situations. }
\label{fig:illumination}
\end{figure}

At the same time, in order to quantitatively demonstrate the illumination changes in VIPL-HR database, we have calculated the mean grey-scale intensity of face area for Situation 1, Situation 4, and Situation 5 in Table~\ref{table:record_situation}. The results are shown in Fig.~\ref{fig:illumination}. We can see that the mean gray-scale intensity varies from 60 to 212, covering complicated illumination variations.

\begin{figure}
\centering
\includegraphics[width=0.6\linewidth, height = 0.20\linewidth]{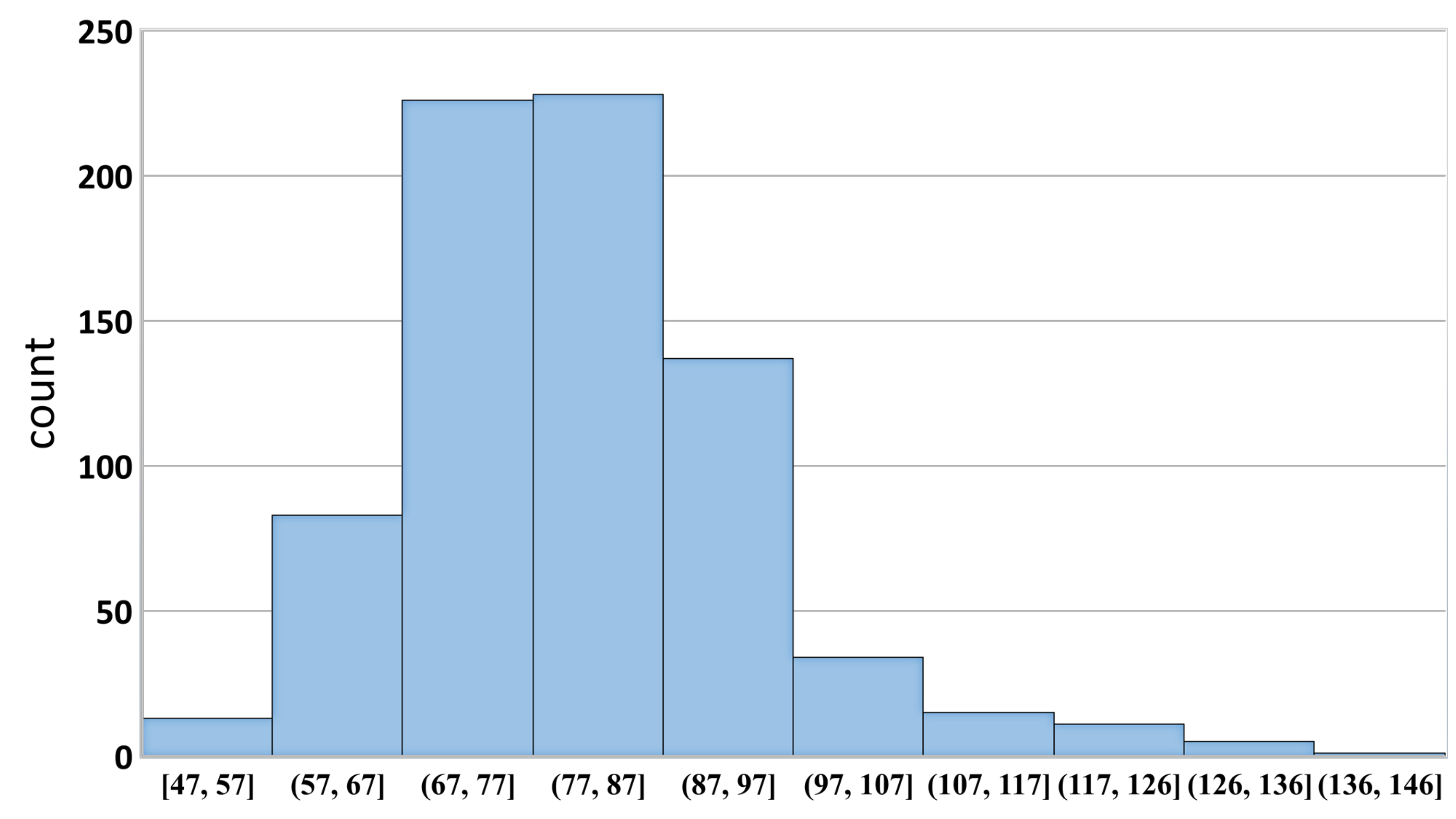}
\caption{ The histogram showing the ground-truth HR distribution. }
\label{fig:gt_HR}
\end{figure}

A histogram of ground-truth HRs is also shown in Fig.~\ref{fig:gt_HR}. We can see that the ground-truth HRs in VIPL-HR vary from 47 bpm to 146bpm, which covers the typical HR range\footnote[3]{\url{https://en.wikipedia.org/wiki/Heart_rate}}. The wide HR distribution in VIPL-HR fills the gap between lab-controlled databases and the HR distribution presenting in daily-life scenes. The relatively large size of VIPL-HR also makes it possible to use deep learning methods to build data-driven HR estimators.

\begin{figure}[t]
      \centering
      \includegraphics[width=0.75\linewidth]{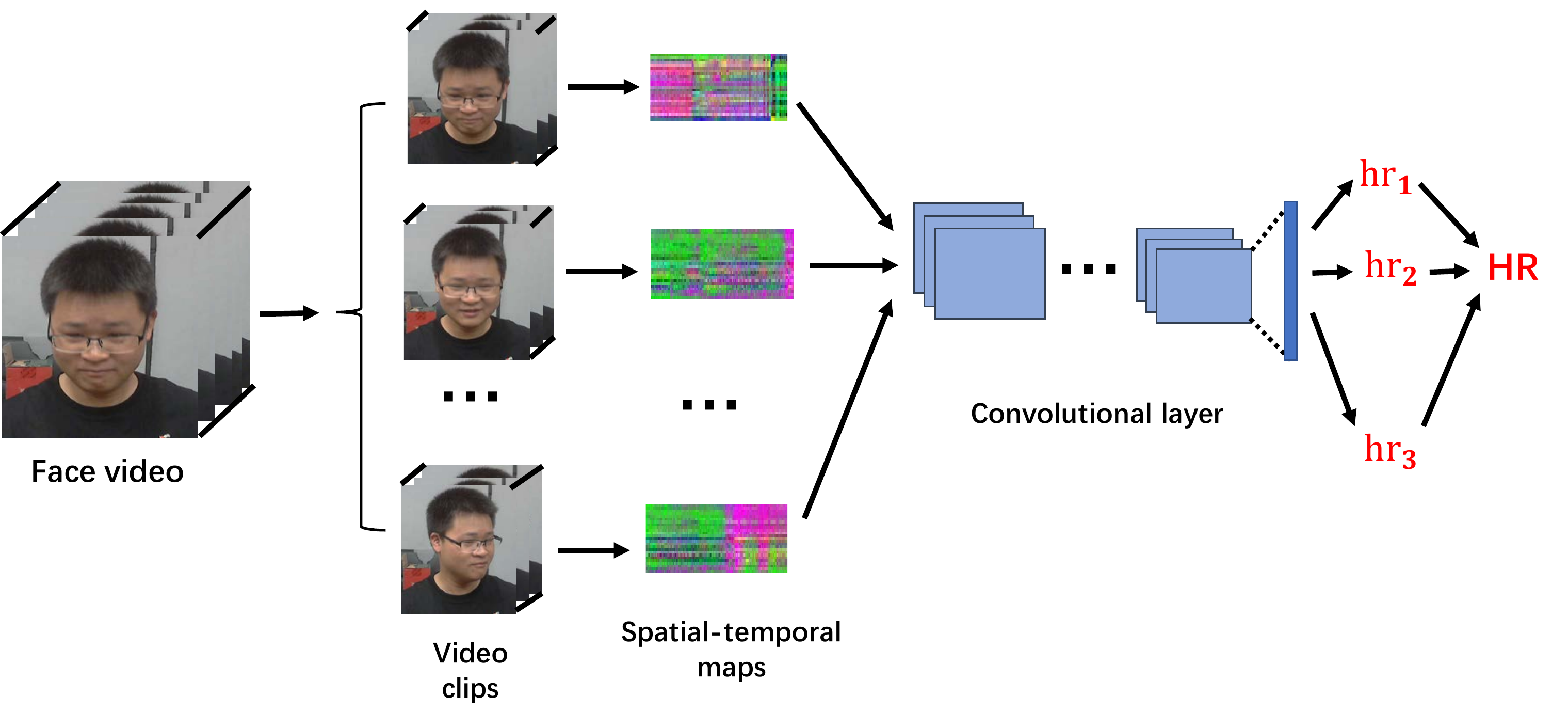}
      \caption{Overview of RhythmNet. Given an input video sequence, we first divide it into multiple short video clips. Then spatial-temporal maps are generated from the aligned face images within each video clip to represent the HR signals, and a CNN model is trained to predict the HR per video clip. Finally, the average HR estimated for a video sequence is computed as the average of all the HRs. }
      \label{fig:OverView}
\end{figure}

\section{Deeply Learned HR Estimator}
\label{Proposed_Approach}

With the less-constrained VIPL-HR database, we are able to build a data-driven HR estimator using deep learning methods. Following the idea of~\cite{niu2018Synrhythm}, we propose a deep HR estimation method, named as RhythmNet. An overview producer of RyhthmNet can be seen in Fig.~\ref{fig:OverView}.

\subsection{Spatial-temporal Maps Generation}

\begin{figure}[t]
      \centering
      \includegraphics[width=0.8\linewidth]{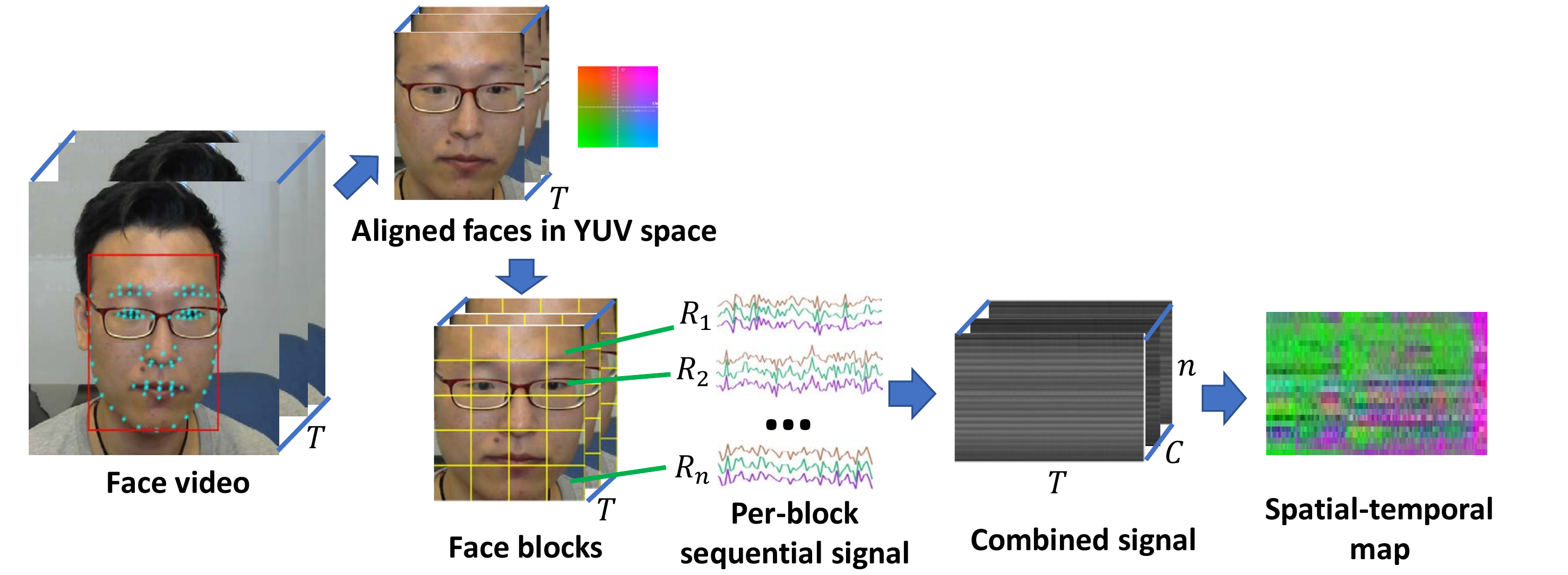}
      \caption{An illustration of the spatial-temporal map generation producer. We first get the face area for each frame and transform the image to YUV color space, then the face area is divided into $n$ blocks, and the average of the pixel values is concatenated into a sequence of $T$ for $C$ channels in each block. The $n$ blocks is directly placed into rows, and we get the final spatial-temporal map with the size of $n\times T \times C$.}
      \label{fig:SignalMap}
\end{figure}

In order to identify the face area within the video frames, we first use the face detector provided by the open source SeetaFace\footnote{\url{https://github.com/seetaface/SeetaFaceEngine}} to get the face
location and 81 facial landmarks (see Fig.~\ref{fig:SignalMap}). Since the facial landmarks detection is able to run at a frame rate of more than 30 fps, we perform face detection and landmarks detection on every frame in order to get consistent ROI localization in a face video sequence. A moving average filter is applied to the 81 landmark points to get more stable landmark localizations.

According to~\cite{Kwon2015ROI}, the most informative facial parts containing the color changes due to heart rhythms are the cheek area and the forehead area. The ROI containing both the cheek and forehead area is determined using the cheek border and chin locations as shown in Fig.~\ref{fig:SignalMap}. Face alignment is firstly performed using the eye center points, and then a bounding box is defined with a width of $w$ (where $w$ is the horizontal distance between the outer cheek border points) and height $1.5*h$ (where $h$ is the vertical distance between chin location and eye center points). Skin segmentation is then applied to the defined ROI to remove the non-face area such as the eye region and the background area.

According to~\cite{niu2018Synrhythm}, a good representation of the HR signals is very important for training a deep HR estimator. Niu et al. directly used the average pixel values of RGB channels as the HR signals representation, which may not be the best choice to represent HR signals. As stated in~\cite{tsouri2015benefits}, alternative color spaces derived from RGB video are beneficial for getting a better HR signal representation.
After testing alternative color spaces, we finally choose the YUV color space for further computing. The color space transform can be formulized as
\begin{align}
\left[
\begin{array}{c}
Y \\
U \\
V \\
\end{array}
\right]
= \left[
\begin{array}{ccc}
0.299 & 0.587 & 0.114\\
-0.169 & -0.331 & 0.5\\
0.5 & -0.419 & -0.081\\
\end{array}
\right]
\left[
\begin{array}{c}
R \\
G \\
B \\
\end{array}
\right] + \left[
\begin{array}{c}
0 \\
128 \\
128 \\
\end{array}
\right]
\end{align}
and the final spatial-temporal map generation producer can be found in Fig.~\ref{fig:SignalMap}.

\subsection{Learning Strategy for HR Measurement}

For each face video, we first divide it into small video clips using a fixed sliding window and then compute the spatial-temporal map, which is used to estimate the HR per video clip. The average HR of the face video is computed as the average of the HRs estimated from each clip. We choose the ResNet18~\cite{he2016deep} as our convolutional neural network (CNN) for learning the mapping from spatial-temporal maps to HRs, which is commonly used in various computer vision tasks. The ResNet18 architecture including four blocks made up of convolutional layers and residual link, one convolutional layer and one fully connected layer for the final classification. The output of the network is one single HR value, which is normalized based on the frame rate of the face video. $L_1$ losses are used for measuring the distance between the predicted HR and ground truth HR.

It is important to leverage the prior knowledge of HR signals during HR estimation. To achieve this purpose, we use the synthetic rhythm data for network pre-training as stated in~\cite{niu2018Synrhythm}. Specifically, our training strategy could be divided into three stages. Firstly, we train our model using the large-scale image database ImageNet~\cite{ILSVRC15}. Then the synthetic spatial-temporal maps are used to guide the network to learn the prior knowledge of mapping a video sequence into an HR value. With this prior knowledge, we can further fine-tune the neural network for the final HR estimation task using the real-life face videos.

Another situation we need to consider is that the face detector may fail in a short time interval, which commonly happens when the subject's head is moving or rotating. The failing of face detection will cause the missing data of HR signals. In order to handle this issue, we randomly mask (set to zero) the spatial-temporal maps along the time dimension to simulate the missing data situation. The masked spatial-temporal maps are found to be useful to train a robust HR estimator against failing of face detection.

\section{Experiments}
\label{experiments}

\subsection{Database, Protocol, and Experimental Settings}
\label{protocol}

In this paper, we use the VIPL-HR database for within-database testing and the MMSE-HR database~\cite{Tulyakov2016Self} for cross-database evaluation. Details about these two databases can be found in Table.~\ref{table:database}. We first perform participant-dependent five-fold cross-validation for within-database testing on the VIPL-HR database. Then, we directly train the RhythmNet on the VIPL-HR database and test it on the MMSE-HR database. After that, the RhythmNet pre-trained on the VIPL-HR is fine-tuned and tested on the MMSE-HR.

Different evaluation metrics have been proposed for remote HR estimation, such as the mean and standard deviation ($HR_{me}$ and $HR_{sd}$) of the HR error, the mean absolute HR error ($HR_{mae}$), the root mean squared HR error ($HR_{rmse}$), the mean error rate percentage ($HR_{mer}$), and Pearson¡¯s correlation coefficients $r$~\cite{li2014remote,Tulyakov2016Self}. In this paper, we also use these evaluation metrics for all the experiments below.

For each video, a sliding window with $300$ frames is used for generating the spatial-temporal maps. We divide the face area into 25 blocks ($5\times 5$ grids). The percentage of masked spatial-temporal maps is $50\%$, and the mask length varies from $10$ frames to $30$ frames. Since the face sizes in the NIR videos are small for the face detector, only 497 NIR videos with detected faces are involved in the experiments. The RhythmNet is implemented using the PyTorch\footnote{\url{https://pytorch.org/}} framework. The Adam solver with an initial learning rate of 0.001 is applied to train the model, and the number of maximum iteration epochs is 50.

\subsection{Within-database Testing}

\subsubsection{Experiments on Color Face Videos.}
We first perform the within-database testing on the VIPL-HR database. We use the state-of-the-art methods (Haan2013~\cite{de2013robust}, Tulyakov2016~\cite{Tulyakov2016Self}, Wang2017~\cite{wang2017algorithmic} and Niu2018~\cite{niu2018Synrhythm}) for comparisons. The results of the individual methods are reported in Table.~\ref{table:vipl_color}.

\setlength{\tabcolsep}{4pt}
\begin{table}[t]
\begin{center}
\caption{The results of HR estimation on color face videos on VIPL-HR database.}
\label{table:vipl_color}
\begin{tabular}{lcccccc}
\hline\noalign{\smallskip}
\hline
\multirow{2}{*}{Method} & $\text{HR}_{me} $ & $\text{HR}_{sd} $ & $\text{HR}_{mae}$ & $\text{HR}_{rmse}$  & \multirow{ 2}{*}{$\text{HR}_{mer}$} & \multirow{ 2}{*}{$r$}\\
&(bpm) &(bpm) &(bpm)&(bpm)\\
\noalign{\smallskip}
\hline
\hline
\noalign{\smallskip}
Haan2013~\cite{de2013robust} & 7.63 & 15.1 & 11.4  & 16.9 & 17.8\% & 0.28 \\
Tulyakov2016~\cite{Tulyakov2016Self} & 10.8 & 18.0 & 15.9 & 21.0 & 26.7\% & 0.11\\
Wang2017~\cite{wang2017algorithmic} & 7.87 & 15.3 & 11.5 & 17.2 & 18.5\% & 0.30\\
\noalign{\smallskip}
\hline
RhythmNet & \textbf{1.02}  & \textbf{8.88} & \textbf{5.79} & \textbf{8.94}   & \textbf{7.38\%}  & \textbf{0.73}\\
\hline
\noalign{\smallskip}
\end{tabular}
\end{center}
\end{table}

From the results, we can see that the proposed method could achieve promising results with an $\text{HR}_{rmse}$ of 8.94bpm, which is a much lower error than other methods. At the same time, it can also be seen that our method achieves a better consistency on the VIPL-HR database with a much higher Pearson¡¯s correlation coefficient $r$ of 0.73.


\setlength{\tabcolsep}{4pt}
\begin{table}[t]
\begin{center}
\caption{The results of HR estimation on NIR face videos on VIPL-HR database.}
\label{table:NIR}
\begin{tabular}{lcccccc}
\hline\noalign{\smallskip}
\hline
\multirow{2}{*}{Method} & $\text{HR}_{me} $ & $\text{HR}_{sd} $ & $\text{HR}_{mae}$ & $\text{HR}_{rmse}$  & \multirow{ 2}{*}{$\text{HR}_{mer}$} & \multirow{ 2}{*}{$r$}\\
&(bpm) &(bpm) &(bpm)&\\
\noalign{\smallskip}
\hline
RhythmNet & 2.87 & 11.08 & 8.50 & 11.44 & 11.0\% & 0.53 \\
\hline
\noalign{\smallskip}
\end{tabular}
\end{center}
\end{table}

\subsubsection{Experiment on NIR Face Videos.}

The experiments on NIR face videos are also conducted using the protocol proposed in Section.~\ref{protocol}. Since the NIR face videos only have one channel, no color space transformation is used, and we get one-channel spatial-temporal maps for the deep HR estimator. Very few methods have been proposed and evaluated on the NIR data, thus we only report the results based on the RhythmNet in Table.~\ref{table:NIR}. 


\setlength{\tabcolsep}{4pt}
\begin{table}
\begin{center}
\caption{The results of average HR estimation per video using different methods on the MMSE-HR database.}
\label{table:mmse_hr}
\begin{tabular}{lccccc}
\hline\noalign{\smallskip}
\hline
\multirow{2}{*}{Method} & $\text{HR}_{me} $ & $\text{HR}_{sd} $  & $\text{HR}_{rmse}$  & \multirow{ 2}{*}{$\text{HR}_{mer}$} & \multirow{ 2}{*}{$r$}\\
&(bpm) &(bpm) &(bpm)&\\
\noalign{\smallskip}
\hline
\hline
\noalign{\smallskip}
Li2014~\cite{li2014remote} & 11.56 & 20.02 & 19.95 & 14.64\% & 0.38 \\
Haan2013~\cite{de2013robust} & 9.41 & 14.08 & 13.97 & 12.22\% & 0.55 \\
Tulyakov2016~\cite{Tulyakov2016Self} & 7.61 &12.24 & 11.37 & 10.84\% & 0.71 \\
\noalign{\smallskip}
\hline
RhythmNet(CrossDB) & -.2.26 &  10.39 &  10.58 &   \textbf{5.35\%}  &   0.69   \\
RhythmNet(Fine-tuned) & \textbf{-1.30} &   \textbf{8.16}  & \textbf{8.22} &  5.54\% & \textbf{0.78} \\
\hline
\noalign{\smallskip}
\end{tabular}
\end{center}
\end{table}

\subsection{Cross-database Testing}

The cross-database experiments are then conducted based on the MMSE-HR database. Specifically, we first train our model on the VIPL-HR database and directly test it on the MMSE-HR database. We also fine-tune the model on MMSE-HR to see whether a finetuning could improve the HR estimation accuracy or not. All the results can be found in Table~\ref{table:mmse_hr}. The baseline methods we use for comparisons are Li2014~\cite{li2014remote}, Haan2013~\cite{de2013robust} and Tulyakov2016~\cite{Tulyakov2016Self}, and their performances are from~\cite{Tulyakov2016Self}.

From the results, we can see that the proposed method could achieve a promising performance with an $HR_{rmse}$ of 10.58 bpm, even when we directly test our VIPL-HR pre-trained model on MMSE-HR. The error rate is further reduced to 8.22 bpm when we fine-tune the pre-trained model on MMSE-HR. Both results of the proposed approach are much better than previous methods.
These results indicate that the variations of illumination, movement, and acquisition device in the VIPL-HR database are helpful to learn an HR estimator which has good generalization ability to unseen scenarios. In addition, the proposed RhythmNet leverages the diverse information contained in VIPL-HR to learn a robust HR estimator.

\section{Conclusion and Further Work}
\label{conculsion}

Remote HR estimation from a face video has wide applications; however, accurate HR estimation from the face in the wild is challenging due to the various variations in less-constrained scenarios. In this paper, we introduce a multi-modality VIPL-HR database for remote heart estimation under less-constrained conditions, such as head movement, illumination change, and camera diversity. We also proposed the RhythmNet, a data-driven heart estimator based on CNN, to perform remote HR estimation. Benefited from the proposed spatial-temporal map and the effective training strategy, our approach achieves promising HR accuracies in both within-database and cross-database testing.

In the future, besides investigating new HR signals representations, we are also going to establish models to leverage the relation between adjacent measurements from the sliding windows. Face dense alignment via 3D modeling~\cite{han20123d} will also be studied to improve ROI alignment. In addition, detailed analysis of individual methods under various recording situations will be provided using the VIPL-HR database. We would also like to apply the proposed approach into the applications of face presentation attack detection~\cite{patel2016secure}, etc.

\section*{Acknowledgement}

This research was supported in part by the National Key R$\&$D Program of China (grant 2017YFA0700804), Natural Science Foundation of China (grants 61390511, 61672496, 61650202), External Cooperation Program of Chinese Academy of Sciences (grant GJHZ1843).

\bibliographystyle{splncs04}
\bibliography{egbib}
\end{document}